\documentclass[letterpaper, 10 pt, conference]{ieeeconf}  %

\IEEEoverridecommandlockouts                              %

\usepackage{amsmath} %
\usepackage{amssymb}  %
\usepackage{graphicx}
\usepackage{subcaption}

\usepackage{tikz}
\usepackage{tikzscale}
\usepackage{pgfplots}
\usetikzlibrary{external}
\usetikzlibrary{graphs,calc,fit,backgrounds,positioning,intersections,fillbetween}

\usepackage{calc}

\usepackage{amsthm}
\newtheorem{definition}{Definition}

\usepackage{cite}

\makeatletter
\let\NAT@parse\undefined
\makeatother
\usepackage{hyperref}

\title{\LARGE \bf
Consecutive Inertia Drift of Autonomous RC Car via Primitive-based Planning and Data-driven Control
}

\author{Yiwen Lu$^{1}$, Bo Yang$^{1}$, Jiayun Li$^{1}$, Yihan Zhou$^{1}$, Hongshuai Chen$^{2}$ and Yilin Mo$^{1}$%
\thanks{This work is in part supported by Tsinghua University - Meituan Joint Institute for Digital Life.}%
\thanks{$^{1}$Y.Lu, B.Yang, J.Li, Y. Zhou, Y.Mo are with Department of Automation, BNRist, Tsinghua University, Beijing, China: {\tt\small \{luyw20, yang-b21, lijiayun22, zhou-yh19\}@mails.tsinghua.edu.cn, ylmo@tsinghua.edu.cn}.
        }%
\thanks{$^{2}$ H.Chen is with the Department of Autonomous Vehicles, Meituan, Beijing, China: {\tt\small chenhongshuai@meituan.com}.}%
}

\begin{document}

\maketitle
\thispagestyle{empty}
\pagestyle{empty}

\begin{abstract}

Inertia drift is an aggressive transitional driving maneuver, which is challenging due to the high nonlinearity of the system and the stringent requirement on control and planning performance.
This paper presents a solution for the consecutive inertia drift of an autonomous RC car based on primitive-based planning and data-driven control.
The planner generates complex paths via the concatenation of path segments called primitives,
and the controller eases the burden on feedback by interpolating between multiple real trajectories with different initial conditions into one near-feasible reference trajectory.
The proposed strategy is capable of drifting through various paths containing consecutive turns, which is validated in both simulation and reality.

\end{abstract}

\section{INTRODUCTION}

Drifting, a driving technique featuring the intentional loss of traction and the maintenance of large slip-slide angles, has been pursued by both professional human racers and autonomous agents~\cite{cutler_autonomous_2016,zhang_drift_2018,goh_toward_2020,cai2020high}. 
Drifting pushes the vehicle to the limit of maneuverability, and hence the study of which may provide insights to full autonomous driving, especially to the precise control of vehicle trajectory when a sudden loss of friction is caused by external conditions like rain or ice~\cite{yang2022hierarchical}.
Furthermore, drift maneuvering is faced with limited control authority in a highly unstable region~\cite{zhang_drift_2018} and a stringent requirement on real-time decision making, which are otherwise ubiquitous challenges in the research of agile robots~\cite{hwangbo2019learning,peng2020learning,levin2019real}. Therefore, control and planning techniques developed for drifting can potentially be of general interest in robotics.

\begin{figure}[!htbp]
    \input{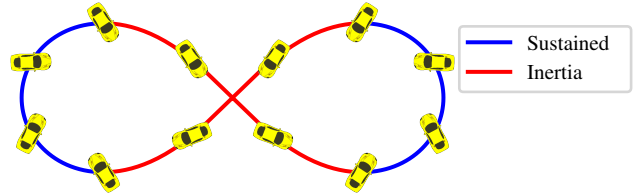}
    \caption{Illustration of a 8-shaped path combining circular and inertia drift.}
    \label{fig:8path}
\end{figure}

Existing research on autonomous drift have mainly focused on either \emph{sustained} or \emph{transient} drift~\cite{zhang_drift_2018}, where the former is concerned with the stabilization near an equilibrium of drifting along a circular path~\cite{yang2022hierarchical,baur2019experimentally,joa2020new}, while the latter refers to temporary drift maneuvers including drift parking and cornering~\cite{zhang_drift_2018,jelavic_autonomous_2017}.
A drift manuever well-known among human enthusiasts that fall outside the above two categories is \emph{inertia drift}, which connects a counter-clockwise drifting path with a clockwise one or vice versa within minimal time and distance, and is useful in racing due to its capability to pass through consecutive turns without significant loss of speed.
The present work combines sustained and inertia drift to accomplish more complex acrobatics than each of the individual maneuvers, e.g., drifting consecutively along the 8-shaped path illustrated in Fig.~\ref{fig:8path}.

The complexity of the drift task considered in this work amounts to various difficulties in control and planning.
On one hand, compared to sustained drift along a unidirectional circular path, inertia drift requires traversing through a series of transient states as opposed to the stabilization near a single equilibrium. Consequently, controllers designed for linearized vehicle models around drift equilibria~\cite{baur2019experimentally} may not apply. In addition, the restricted availability of trajectory data in each particular region of the state space poses challenge to the online system identification that facilitates model-based control~\cite{joa2020new}.
On the other hand, compared to a transient drift, the combination of large sideslip angles at both the beginning and the end of an inertia drift trajectory demands higher controller performance.
To be capable of following the path consecutively rather than performing only a one-shot maneuver, the agent also needs a planner that chooses the accurate timings for switching between inertia and sustained drift modes in real time.

The present work addresses the aforementioned challenges through primitive-based planning and data-driven control.
By decomposing the drifting path into motion primitives including sustained and inertia drift trajectory segments, the planning problem is reduced to selecting a motion primitive stored offline at each moment based on simple geometry. The primitive-based planner is versatile in the sense that it can be applied to various tasks as long as the desired path comprises known primitives. It is also computationally light compared to sampling-~\cite{karaman2011sampling} or optimization-based~\cite{kalakrishnan2011stomp} planning methods, and plans onboard at 100Hz.
Meanwhile, through the combination of only a handful real trajectories stored offline, the data-driven controller can generate near-feasible reference readily tracked by simple feedback, without relying on explicit system identification.
The proposed strategy is verified on both a high-fidelity simulation platform and a real 1/10-scale RC car.

The contributions of this work can be summarized as follows:
\begin{enumerate}
    \item This work achieves the autonomous control of \emph{inertia drift}, a challenging maneuver that has not been explicitly studied in the literature.
    \item Various complex maneuvering tasks unaccomplished by previous methods~\cite{goh_toward_2020,domberg_deep_2022}, including drifting along an 8-shaped path, a three-circle path, and an ``Olympic Rings'' path (refer to Fig.~\ref{fig:traj}) within a confined space, are tackled by the proposed method in a unified manner.
    \item The proposed method is physically verified by an RC car  with onboard planning and control (refer to video attachment), demonstrating the promise of aggressive real-world maneuvering.
\end{enumerate}

The rest of the manuscript is organized as follows: Section~\ref{sec:related} gives a brief review of existing research on autonomous drift, as well as on primitive-based motion planning and data-driven control.
Section~\ref{sec:platform} introduces the hardware and simulation platforms for this work, summarizes the modeling process, and provides a problem statement.
Section~\ref{sec:method} describes the proposed planning and control strategy.
Section~\ref{sec:result} presents the result of the strategy on various drift maneuvering tasks, both in simulation and in reality.
Finally, Section~\ref{sec:conclusion} gives concluding remarks.

\section{RELATED WORKS}
\label{sec:related}

\subsection{Autonomous drift}

\paragraph{Planning-and-control-based methods}
As mentioned in the introduction, drifting tasks can be roughly categorized into sustained and transient drift~\cite{zhang_drift_2018}, which, from the perspective of control, correspond to stabilization and tracking problems respectively.

Among works on sustained drift, Baur et al.~\cite{baur2019experimentally} applied an LQR controller on a linearized model near an equilibrium, Joa et al.~\cite{joa2020new} proposed a force-based control design with online system identification, and Yang et al.~\cite{yang2022hierarchical} designed a curvature inference and feedback scheme for the stabilization of slowly-varying drift equilibria.

Among works on transient drift, typically studied maneuvers include drift parking and drift cornering. For drift parking, Jelavic et al.~\cite{jelavic_autonomous_2017} proposed a rule-based planner and a switched MPC / feedforward-feedback controller. For drift cornering, Zhang et al.~\cite{zhang_drift_2018} proposed a hybrid RRT and rule-based planner as well as a mixed open-and-closed-loop controller, and Malmir et al.~\cite{malmir2018model} used nonlinear optimal control for offline reference trajectory generation and LTV-MPC for online tracking.

Unified solutions to sustained and transient drift have also been attempted. Goh et al.~\cite{goh_toward_2020} designed a controller for drifting along general paths based on a trajectory-dependent curvilinear coordinate system and nonlinear model inversion, based on which Goel et al.~\cite{goel_opening_2020} added brake control for increased flexibility. However, the practicality of the above works are limited by their reliance on the accurate parameterization of reference trajectory and modeling of vehicle dynamics.
Bellegarda et al.~\cite{bellegarda2020versatile} performed trajectory optimization for a variety of maneuvers based on a Coulomb tire model, whose solution is nevertheless open-loop.

In contrast to the aforementioned works, we explicitly consider the planning and control involving inertia drift, whose capability to tackle a variety of previously unaccomplished maneuvering tasks is demonstrated on a physical RC car.

\paragraph{Learning-based methods} Drift maneuvering has also been attempted using reinforcement learning. Culter et al.~\cite{cutler_autonomous_2016} applied a model-based reinforcement learning algorithm called PILCO to stabilize a particular drift equilibrium. Cai et al.~\cite{cai2020high} achieved high-speed simulated racing by training the agent to track expert trajectories with model-free Deep Reinforcement Learning (DRL), specifically the SAC algorithm. Domberg et al.~\cite{domberg_deep_2022} attempted an end-to-end DRL solution (specifically the PPO algorithm) to the 8-shaped drift maneuver similar to the one presented in the current work, as well as the multi-circle trajectory in~\cite{goh_toward_2020}, which nevertheless relies on handcrafted reward functions and undergoes considerable performance degradation when transferred from simulation to reality.
By contrast, the proposed data-driven control method is based on the interpolation of known trajectories instead of the learning of parameterized function approximators such as deep neural networks, and therefore enjoys high data efficiency and ease of deployment on physical cars.

\subsection{Primitive-based motion planning}

The idea of using motion primitives to relieve the planner of the computational burden stemming from system dynamics and nonholonomic constraints was originated and developed by Frazzoli et al.~\cite{frazzoli1999hybrid,frazzoli2002real}.
The aforementioned framework has been applied to quadrotor~\cite{vukosavljev2019modular} and fixed-wing~\cite{levin2019real} aircraft, legged robots~\cite{apostolopoulos2017energy} and autonomous driving~\cite{wang2018predictive}.
To our knowledge, the application of primitive-based planning to aggressive driving scenarios has not been widely studied, due in part to its sensitivity to model mismatch, exacerbated by tire saturation, which the present work addresses via the coupling of primitive-based planning and data-driven control.

\subsection{Data-driven control}

The key idea of data-driven control is describing and controlling a system in terms of measured data instead of learning a parametric system representation, which has long been studied in the control literature~\cite{willems2005note,de2019formulas}. The aforementioned works tend to be theoretically inclined.
Recent attempts of developing data-driven robotics control methods include Chen et al.~\cite{chen2019hardware}, which proposed a hardware-in-the-loop robot arm predictive controller using measured data instead of exact model for future prediction.
The current work proposes a novel application of the idea of data-driven control to the problem of autonomous car maneuvering.

\section{PRELIMINARIES}
\label{sec:platform}

\subsection{Target platform}
\label{sec:platform1}

\begin{figure}[!htbp]
    \centering
    \includegraphics[width=0.5\columnwidth]{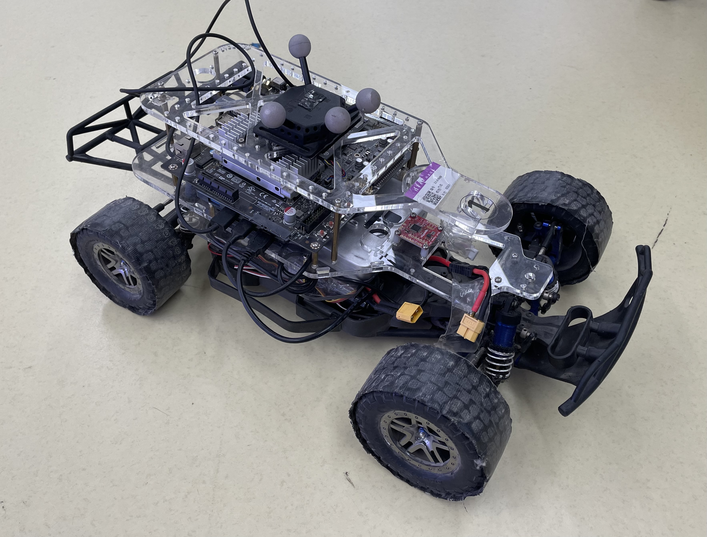}
    \caption{The RC car platform}
    \label{fig:car}
\end{figure}

The target platform of the proposed strategy is a 1/10-scale, four-wheel-driven RC car with locked differential shown in Fig.~\ref{fig:car}, inspired by MIT Racecar~\cite{karaman2017project}, Berkeley BARC~\cite{rosolia2019learning} and F1TENTH~\cite{o2020f1tenth} projects.
The car is equipped with an NVIDIA Jetson TX2 single board computer which runs the planner and controller onboard, and controls the electric motor of the car through a VESC motor controller.
High-accuracy pose information is provided by a motion capture system, based on which the (angular) velocities and accelerations are estimated using a Kalman filter.

For strategy design and verification, a self-developed mixed-fidelity simulation platform is used. The platform is implemented using the \texttt{DifferentialEquations}~\cite{rackauckas2017differentialequations} package in Julia, and supports both a bicycle model with Pacejka tire~\cite{karaman2011sampling}, which is used for strategy design, and a higher-fidelity four-wheel model~\cite{milani2022vehicle} with actuator delay, which is used for strategy verification.

\subsection{Modeling}

\begin{figure}[!htbp]
    \centering
    \hspace{-1.5cm}
    \resizebox{\columnwidth}{!}{
       \includegraphics{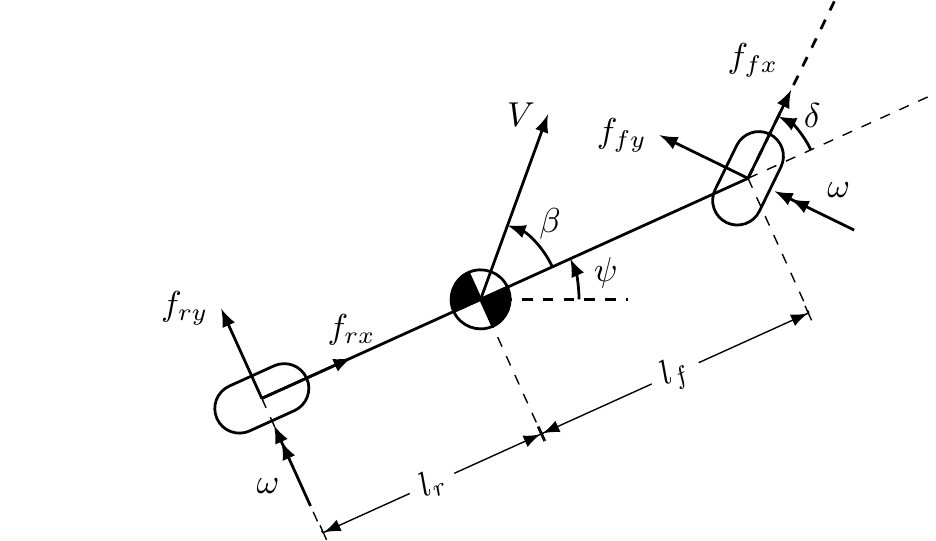}
    }
    \caption{Illustration of bicycle model}
    \label{fig:bicycle}
\end{figure}

The proposed strategy is based on the bicycle model similar to the one used in~\cite{yang2022hierarchical,lu_two-timescale_2021}, illustrated in Fig.~\ref{fig:bicycle}. The state and input variables of this bicycle model are:
\begin{equation}
    \mathbf{X}=(x, y, \psi, \dot{x}, \dot{y}, \dot{\psi}), \mathbf{U}=(\delta, \omega),
    \label{eq:xu}
\end{equation}
where $x,y$ are the coordinate of the center of mass of the car in the world frame, $\psi$ is the yaw angle, $\delta$ is the front wheel steering angle, and $\omega$ is the rotational speed of both wheels.

Consider the following variables that describe the motion of the car:
\begin{align}
    r &= \dot \psi && \text{(yaw rate),}\\
    \beta &= \operatorname{atan2}(\dot y, \dot x) - \psi && \text{(sideslip angle),}\label{eq:beta}\\
    V &= \sqrt{\dot{x} ^ 2 + \dot{y} ^ 2} && \text{(speed)}.
\end{align}
Assuming that \( V > 0 \), similarly to~\cite{goh_toward_2020}, one can derive a reduced-order model w.r.t.  \( (r, \beta, V) \) as follows:
\begin{equation}
    \begin{aligned}
    \dot{r} &=\frac{l_{f}\left(f_{f y} c_{\delta}+f_{f x} s_{\delta}\right)-l_r f_{r y}}{I_{z}}, \\
    \dot{\beta} &=\frac{f_{f y} c_{\delta-\beta}+f_{f x} s_{\delta-\beta}+f_{r y} c_{\beta}-f_{r x} s_{\beta}}{m V}-r, \\
    \dot{V} &=\frac{-f_{f y} s_{\delta-\beta}+f_{f x} c_{\delta-\beta}+f_{r y} s_{\beta}+f_{r x} c_{\beta}}{m},
    \end{aligned}
    \label{eq:rbV}
\end{equation}
where $c_\cdot, s_\cdot$ stand for $\cos(\cdot), \sin(\cdot)$ respectively, $m$ is the mass of the car, $I_z$ is the moment of inertia of the car about the vertical axis, $l_f,l_r$ are the distances from the center of mass to the front and rear axles respectively, and $f_{fx}, f_{fy}, f_{rx}, f_{ry}$ are the components of friction forces as exemplified in Fig.~\ref{fig:bicycle}. The input of this reduced-order model is $(\delta, f_{fx}, f_{fy}, f_{rx}, f_{ry})$, which can be derived from the actual input $(\omega, \delta)$ through the Pacejka tire model.

It is worth noticing that given $(r,\beta,V)$, the motion of the pose $(x,y,\psi)$ can be determined by the following purely kinematic equation:
\begin{equation}
    \dot{x}=V \cos (\beta+\psi), \;
    \dot{y}=V \sin (\beta+\psi), \;
    \dot{\psi}=r,
    \label{eq:xypsi}
\end{equation}
which legitimates~\eqref{eq:rbV} as a rotation- and translation-invariant description of the motion of the car, and lays the foundation for the definition of motion primitive. We shall refer to $(r,\beta,V)$ as a ``reduced state'' hereafter.

\subsection{Problem statement}

The aim of this work is to use steering and speed commands (which correspond to the control inputs \( \delta, \omega \) in~\eqref{eq:xu}) adhering to actuation limits to perform autonomous maneuvers that interleave sustained drift with inertia drift, within a confined space. The two types of drift are both characterized by a nontrivial sideslip angle \( \beta \) (c.f.~\eqref{eq:beta}), which are described respectively as follows:
\begin{itemize}
    \item Sustained drift: maintaining an approximately constant sideslip angle, e.g., \( \beta \approx 1 \,\mathrm{rad} \) for a clockwise path or \( \beta \approx -1 \,\mathrm{rad} \) for a counterclockwise path, while traveling along a prescribed circle. Let the desired sideslip angle be \( \beta_{ \mathrm{ref}} \), and the prescribed circle be centered at \( (x_c, y_c) \) with radius \( R \), then we define the error of sustained drift to be
    \begin{align}
        & \begin{aligned}
        & e = (e_{ \mathrm{slip}}, e_{ \mathrm{pos}}, e_{ \mathrm{dir}}), \text{   where} \\
        & e_{ \mathrm{slip}} = \beta - \beta_{ \mathrm{ref}},  \\
        & e_{ \mathrm{pos}} = \sqrt{\left(x-x_c\right)^2+\left(y-y_c\right)^2}-R,
        \end{aligned} \label{eq:error}\\
        & e_{ \mathrm{dir}} = \psi + \beta -  \mathrm{atan2}(y_c - y, x_c - x) + \frac{ \mathrm{sgn}(\beta)\pi}{2}. \nonumber
    \end{align}
    The definitions of the position error \( e_{ \mathrm{pos}} \) and the direction error \( e_{ \mathrm{dir}} \) are illustrated in Fig.~\ref{fig:error}.
    \begin{figure}[!htbp]
        \centering
        \includegraphics{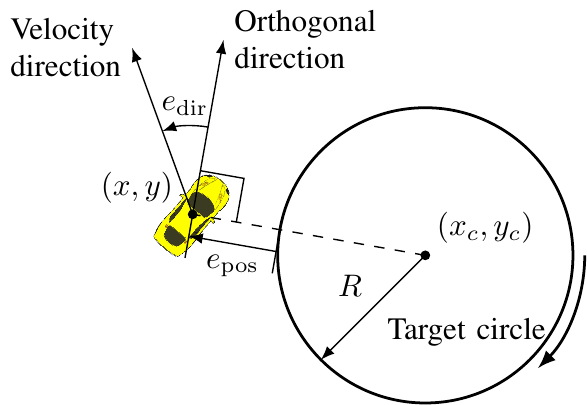}
        \caption{Illustration of the error terms \( e_{ \mathrm{pos}} \) and \( e_{ \mathrm{dir}} \): \( e_{ \mathrm{pos}} \) is the signed distance between the car and the target circle, and \( e_{ \mathrm{dir}} \) is the difference between the velocity direction \( \psi + \beta \) and the desired direction \( \mathrm{atan2}(y_c - y, x_c - x) - \mathrm{sgn}(\beta)(\pi/2) \), which is orthogonal to the line passing the car and the center of the target circle.}
        \label{fig:error}
    \end{figure}
    \item Inertia drift: rapidly altering the sign of the sideslip angle, e.g., from \( \beta \approx -1 \, \mathrm{rad} \) to \( \beta \approx 1\, \mathrm{rad} \) or vice versa, while traveling along an S-shaped path. This is the transitional maneuver between two stages of sustained drift in opposite directions. Therefore, instead of defining a dedicated error function for inertia drift, we consider an inertia drift maneuver to be successful if it creates a desirable initial condition for the subsequent sustained drift maneuver.
\end{itemize}
The overall objective is to track a path defined by multiple target circles, using sustained drift when traveling along each circle and inertia drift when moving between circles, with the error defined in~\eqref{eq:error} bounded across different stages of sustained drift.
Instances of target paths include the 8-shaped path shown in Fig.~\ref{fig:8path} and other more complex paths comprising three or more circles show in Fig.~\ref{fig:traj}.
These are common training tasks for drift enthusiasts, and can therefore serve as benchmarks of autonomous car maneuvering in challenging scenarios.

\begin{figure*}[!htbp]
    \centering
        \includegraphics{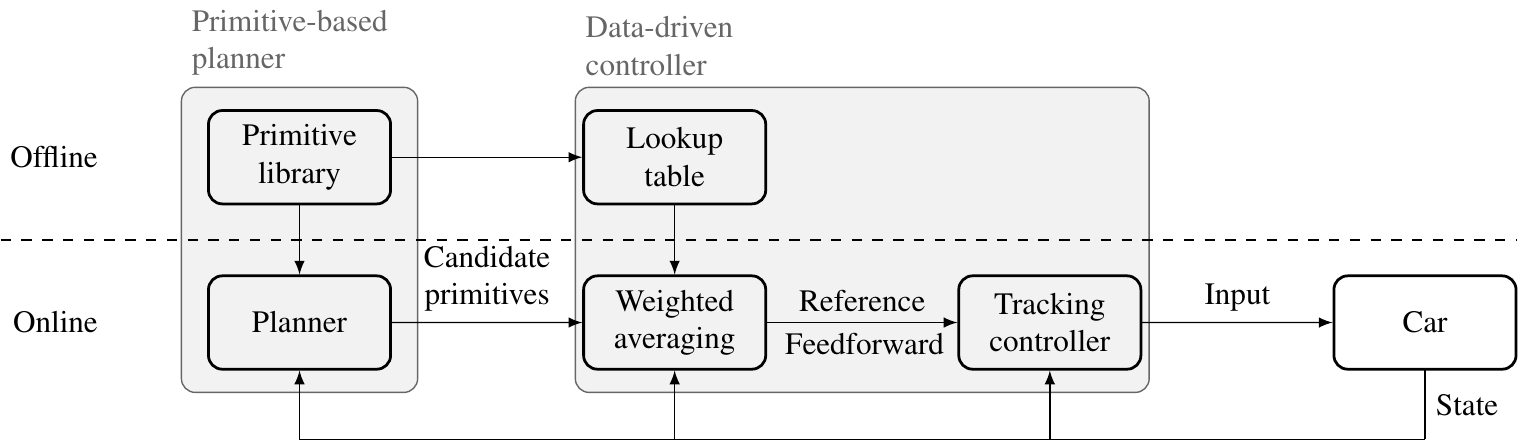}
    \caption{Overview of the proposed planning and control strategy}
    \label{fig:overview}
\end{figure*}

\section{METHOD}
\label{sec:method}

This section is intended to resolve:
\begin{enumerate}
    \item \emph{Planning}: How to select the timings of switching between sustained and inertia drift according to the task specification?
    \item \emph{Control}: How to track sustained and inertia drift paths respectively?
\end{enumerate}

For the control of sustained drift, curvature feedback for controlling the path radius, along with a circumnavigation rule~\cite{dong2020target} for stabilizing the path center, has proved effective and robust~\cite{yang2022hierarchical}, and is therefore adopted for this work. Interested readers are referred to~\cite{yang2022hierarchical} for the technical details.
The remainder of this section is dedicated to the high-level planning and the control of inertia drift.

An overview of the proposed strategy is shown in Fig.~\ref{fig:overview}.
The primitive-based planner selects the timings of switching by referring to an offline-generated library of inertia drift primitives.
Once it decides to enter inertia drift, it passes a subset of the library, known as candidate primitives, to the data-driven controller.
The latter determines a near-feasible trajectory by taking the weighted average of the candidate primitives utilizing a pre-computed lookup table, and tracks the near-feasible trajectory with a feedforward / feedback controller.
The details of planning and control are described respectively below.

\subsection{Primitive-based planning}

To keep complex maneuvers tractable, one can decompose a task into relatively simple motion primitives. In particular, we focus on primitives of drift. Formally, a drift primitive can be defined as follows:

\begin{definition}
    A drift primitive is defined as the following 5-tuple:
    \begin{equation}
        \begin{aligned}
            & \mathcal{P} = \left(T,\tau, \Delta x_b, \Delta y_b, \Delta \psi\right), \\
            & \text{with }\tau = \left\{\left(r(t),\beta(t),V(t),\omega(t),\delta(t)\right)\right\}_{t=1}^T,
        \end{aligned}
        \label{eq:prim}
    \end{equation}
    where $T\in \mathbb{N}^*$ is the number of time steps, $\tau$ is a trajectory of the quantities \( r, \beta, V, \omega, \delta \) defined in~\eqref{eq:xu} and~\eqref{eq:rbV} sampled over time, $\Delta x_b, \Delta y_b$ are the difference between terminal and initial $x,y$ coordinates as measured in the initial body frame, and $\Delta \psi$ is the difference between terminal and initial yaw angles.
\end{definition}

Note that the representation of drift primitive in~\eqref{eq:prim} may seem redundant, as the quantities \( \Delta x_b, \Delta y_b, \Delta \psi \) may all be inferred from the trajectory \( \tau \). However, as it will become clear later, the storage of these quantities allows real-time planning without the need for access to the entire trajectory. Therefore, it is an essential feature of the proposed primitive-based planning method.

Complex drift maneuvers can be achieved through the concatenation of sustained and inertia drift primitives. The rest of this subsection is devoted to explaining how to generate inertia drift primitives offline and how to perform the concatenation online.

\subsubsection{Generation of inertia drift primitives}

Generating an inertia drift primitive is in essence finding a dynamically feasible path from an initial reduced state $(r_0, \beta_0, V_0)$ to a desired terminal reduced state $(r_{\mathrm{des}}, \beta_{\mathrm{des}}, V_{\mathrm{des}})$, where $r_0$ (resp. $\beta_0$) and $r_{\mathrm{des}}$ (resp. $\beta_{ \mathrm{des}}$) have opposite signs. Therefore, given $(r_0, \beta_0, V_0)$ and $(r_{\mathrm{des}}, \beta_{\mathrm{des}}, V_{\mathrm{des}})$, a feasible primitive for the bicycle model can be generated by solving the following trajectory optimization problem and sampling the resulting trajectory:
\begin{align}
    & \min_{t_f, r,\beta,V,\omega,\delta}\; J = \int_{0}^{t_f} (r_\omega \omega(t)^2 + r_\delta \delta(t)^2) \mathrm{d}t, \nonumber \\
    & \text{s.t. } r(0) = r_0, \beta(0) = \beta_0, V(0) = V_0, \nonumber \\
    & \quad \| (r(t_f),\beta(t_f),V(t_f)) - (r_{\mathrm{des}}, \beta_{\mathrm{des}}, V_{\mathrm{des}}) \| \leq \epsilon, \label{eq:opt}\\
    & \quad 0 \leq \omega(t) \leq \omega_{\max}, -\delta_{\max} \leq \delta(t) \leq \delta_{\max}, \,\forall t\in [0,t_f], \nonumber \\
    & \quad \text{and dynamics constraint~\eqref{eq:rbV},} \nonumber
\end{align}
where $r_\omega, r_\delta$ are weight coefficients for penalizing large inputs, $\epsilon > 0$ a tolerance threshold for inexact terminal state, \( \omega_{\max}, \delta_{\max} \) are the actuation limits, and the forces \( f_{fx}, f_{fy}, f_{rx}, f_{ry} \) in the dynamics constraint~\eqref{eq:rbV} are posed as functions of the inputs \( (\omega(t), \delta(t)) \) via the Pacejka tire model.
With the total time \( t_f \) fixed, the above trajectory optimization problem can be solved by iLQR, and the optimal \( t_f \) can be search via bisection.

The primitive obtained from solving~\eqref{eq:opt}, however, may be infeasible for the target platform, i.e., either high-fidelity simulation or real car, due to modeling errors caused by the simplifications in the bicycle model and parameter mismatch. Therefore, during offline calibration, each primitive generated using the bicycle model is fed to the target platform as a reference trajectory and tracked by a feedforward / feedback controller. The resulting \emph{real} trajectory, once passing a check that the terminal $(r(t_f),\beta(t_f),V(t_f))$ is not too far from the desired $(r_{\mathrm{des}}, \beta_{\mathrm{des}}, V_{\mathrm{des}})$, is stored into a primitive library for the online planner and controller to query.

\subsubsection{Planner logic}

The goal of online planning is concatenating motion primitives with appropriate timings, specifically interleaving sustained drift with inertia drift. The logic of the planner can be represented by the finite state machine in Fig.~\ref{fig:fsm}. The conditions that cause the planner to switch from one primitive to another are explained as follows:

\begin{figure}[!htbp]
    \centering
    \includegraphics{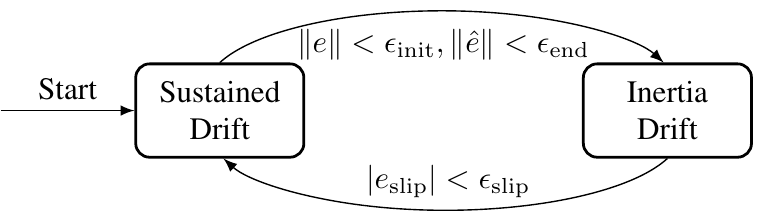}
    \caption{Planner logic: \( e \) is the current error against the current target circle; \( \hat{e} \) is the predicted terminal error against the next target circle if inertia drift is applied now; \( e_{ \mathrm{slip}} \) is the error in the sideslip angle; $\epsilon_{ \mathrm{init}}, \epsilon_{ \mathrm{end}}, \epsilon_{\text{slip}}$  are design constants for error tolerance. Refer to~\eqref{eq:error} and Fig.~\ref{fig:error} for the definition of error.}
    \label{fig:fsm}
\end{figure}

\paragraph{From sustained drift to inertia drift}
Assume that the current pose of the car is $(x,y,\psi)$, the current reduced state is $(r, \beta, V)$, and the error defined in~\eqref{eq:error} against the current target circle for sustained drift is smaller than a threshold. For a particular drift primitive $\mathcal{P}$ as specified in~\eqref{eq:prim}, the planner attempts to ``place'' it in the world frame applying $(\Delta x_b, \Delta y_b, \Delta \psi)$ to $(x,y,\psi)$ as follows:
\begin{align*}
    & \begin{bmatrix}
        \Delta x \\
        \Delta y
    \end{bmatrix} =
    \begin{bmatrix}
        \cos\psi & -\sin\psi \\
        \sin\psi & \cos\psi
    \end{bmatrix}
    \begin{bmatrix}
        \Delta x_b \\
        \Delta y_b
    \end{bmatrix},\\
    & (\hat{x}, \hat{y}, \hat{\psi}) = (x + \Delta x, y + \Delta y, \psi + \Delta \psi),
\end{align*}
where $(\hat{x}, \hat{y}, \hat{\psi})$ stands for the predicted terminal pose if the planner switches to $\mathcal{P}$ immediately. If $(\hat{x}, \hat{y}, \hat{\psi})$, along with $(r(T), \beta(T), V(T))$ stored in \( \mathcal{P} \), fits the target circular path from the task specification, then it can be said it is appropriate timing to switch to $\mathcal{P}$. In particular, the ``fitness'' to a target circle can be measured by the error defined in~\eqref{eq:error}, and switching happens when the error is small enough.

So far, it remains undiscussed how to select a primitive $\mathcal{P}$ from the primitive library. Considering the actual initial reduced state $(r, \beta, V)$ may not match  \( (r(1), \beta(1), V(1)) \) of any primitive from the pre-stored library, it is necessary to interpolate from multiple primitives. Such interpolation is closely related to data-driven control, and the detailed discussion of that aspect shall be deferred to Subsection~\ref{sec:control}.

\paragraph{From inertia drift to sustained drift} As revealed by previous research~\cite{yang2022hierarchical,baur2019experimentally}, a key to successfully maintaining sustained drift is a large initial sideslip angle $\beta$. Therefore, as shown in Fig.~\ref{fig:fsm}, sustained drift can take over from inertial drift when $\beta$ is close to the target value of the next target circle.

\subsubsection{On boundedness of trajectory}

Assuming that both sustained and inertia drift trajectories are tracked by stabilizing controllers, it can be shown via a Lyapunov argument that under the aforementioned planner, the trajectory error (defined in~\eqref{eq:error}) is always bounded (proof omitted due to space limit).

Now let us check the validity of the above prerequisites of boundedness:
while the stability of closed-loop sustained drift has been proved in~\cite{joa2020new}, the stability of closed-loop inertia drift, which is an extremely challenging nonlinear tracking problem, has not been rigorously established.
Therefore, we resort to data-driven control described in the next subsection for generating near-feasible reference trajectories and hence easing the burden of tracking.
The overall closed-loop system integrating planning and control empirically demonstrates stability, as will be shown in Section~\ref{sec:result}.

\subsection{Data-driven control}
\label{sec:control}

Data-driven control is concerned with generating and tracking near-feasible trajectories under particular initial conditions using collected data.
This subsection is devoted to developing the data-driven control technique for inertia drift primitives.

\subsubsection{Posing data-driven control as a weighted average of primitives}
The goal of data-driven control in this work is to find a near-feasible primitive for the current initial reduced state $(r,\beta,V)$, given multiple candidate primitives with different initial reduced states from the library. Let $\mathcal{P}_1,\ldots, \mathcal{P}_n$ be $n$ candidate primitives. Intuitively, a near-feasible primitive starting from $(r,\beta,V)$ can be obtained by the weighted average of $\mathcal{P}_1,\ldots, \mathcal{P}_n$, with the weight vector $\lambda \in \mathbb{R}^n$ determined by:
\begin{align}
    & \min \| \lambda \|_2^2 \label{eq:interp_obj} \\
    & \text{s.t. } \sum_{i=1} ^ n \lambda_i \cdot (r_{i}(1), \beta_{i}(1), V_{i}(1)) = (r, \beta, V), \label{eq:interp_init}\\
    & \quad\;\, \sum_{i=1}^n \lambda_i = 1,\; \lambda_i \geq 0, i = 1,\ldots, n, \label{eq:interp_norm}
\end{align}
where $(r_{i}(i), \beta_{i}(1), V_{i}(1))$ is $(r(1),\beta(1),V(1))$ from $\mathcal{P}_i$, constraint~\eqref{eq:interp_init} makes sure the initial reduced state of the new primitive matches $(r_0, \beta_0, V_0)$, and constraint~\eqref{eq:interp_norm} guarantees the weights are normalized and nonnegative. The objective~\eqref{eq:interp_obj}, i.e., minimum 2-norm, promotes an even distribution of weights. This is based on the intuition that primitive data collected from a real car may be corrupted by random noise, which can be alleviated by averaging multiple trajectories.

It is assumed that the optimization problem~\eqref{eq:interp_obj}-\eqref{eq:interp_norm} is feasible, i.e., $(r, \beta, V)$ belongs to the convex hull spanned by $\{(r_{i}(1), \beta_{i}(1), V_{i}(1))\}_{i=1}^n$, which can be ensured when the candidate primitives have sufficiently diverse initial reduced states.

\subsubsection{Weighted average of primitives via soft dynamic time warping}

Given the vector $\lambda$ determined by~\eqref{eq:interp_obj}-\eqref{eq:interp_norm}, defining the average of $\mathcal{P}_1,\ldots, \mathcal{P}_n$ weighted by $\lambda$ is still highly nontrivial: by~\eqref{eq:prim}, each primitive $\mathcal{P}_i$ contains a trajectory $\tau_i = \left\{\left(r_{i}(t),\beta_{i}(t),V_{i}(t),\omega_{i}(t),\delta_{i}(t)\right)\right\}_{t=1}^{T_i}$, whose length $T_i$ may vary across primitives, and therefore naive methods for weighted averaging do not apply.
To address the aforementioned problem, one can cast the weighted averaging of trajectories $\tau_1,\ldots,\tau_n$ with weight vector $\lambda$ into the following optimization problem:
\begin{equation}
    \min_{\bar{\tau}} \sum_{i=1}^n \lambda_i d(\bar{\tau}, \tau_i),
    \label{eq:barycenter}
\end{equation}
where $d$ is a generalized distance metric function between trajectories.
In particular, we choose \( d \) to be the \( \operatorname{dtw}_\gamma \) metric~\cite{cuturi2017soft}, a log-sum-exp approximation of the minimum distance among all possible alignments between two temporal sequences that may vary in length and speed. Readers are referred to~\cite{cuturi2017soft} for the exact definition of the \( \operatorname{dtw}_\gamma \) metric. We choose \( \operatorname{dtw}_\gamma \) over other metrics for sequences~\cite{sakoe1978dynamic,cuturi2007kernel} because it is differentiable (where the parameter \( \gamma \) controls the smoothness), and therefore allows for finding a local minimum of~\eqref{eq:barycenter} using gradient-based optimizers.
Based on the above discussions, the weighted average of primitives can be defined as follows:

\begin{definition}
    For drift primitives $\mathcal{P}_i = (T_i, \allowbreak \tau_i, \allowbreak \Delta x_{bi}, \allowbreak \Delta y_{bi}, \allowbreak \Delta \psi_i)\,(i=1,\ldots,n)$, a weight vector $\lambda \in \mathbb{R}^n$ satisfying~\eqref{eq:interp_norm}, and a scalar $\gamma > 0$, the weighted average of $ \mathcal{P}_1,\allowbreak \ldots, \allowbreak \mathcal{P}_n$ weighted by $\lambda$ with smoothness $\gamma$ is defined as:
    \begin{equation}
        \bar{\mathcal{P}} = (\bar{T}, \bar{\tau}, \overline{\Delta x_b}, \overline{\Delta y_b}, \overline{\Delta \psi}),
        \label{eq:avg_prim}
    \end{equation}
    where $\bar{T} = \sum_{i=1}^n \lambda_i T_i$, $\bar{\tau}$ is the solution of~\eqref{eq:barycenter} with $d = \operatorname{dtw}_\gamma$ defined in~\cite{cuturi2017soft}, $\overline{\Delta x_b}=\sum_{i=1}^n \lambda_i \Delta x_{bi}$, $\overline{\Delta y_b}=\sum_{i=1}^n \lambda_i \Delta y_{bi}$ and $\overline{\Delta \psi}=\sum_{i=1}^n \lambda_i \Delta \psi$.
    \label{def:weighted_avg}
\end{definition}

\begin{figure}[!htbp]
    \hspace{0.2cm}\begin{subfigure}{0.5\columnwidth}
        \centering
        \hspace{-0.5cm}\includegraphics{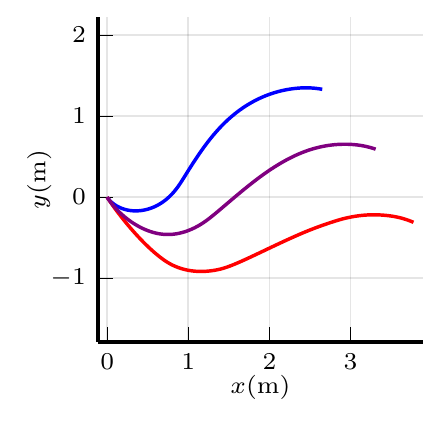}\vspace{-0.5cm}\caption{Trajectory}
    \end{subfigure}\begin{subfigure}{0.5\columnwidth}
        \centering
        \hspace{-0.5cm}\includegraphics{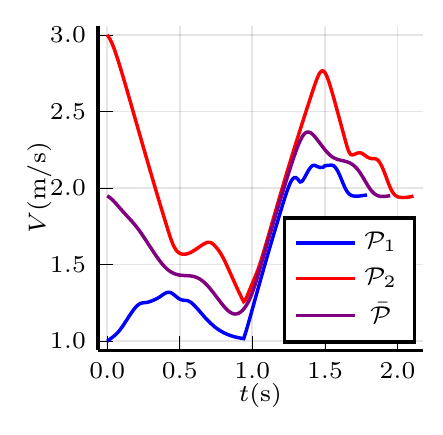}\vspace{-0.5cm}\caption{Velocity profile}
    \end{subfigure}
    \caption{Demonstration of averaging drift primitives via the \( \operatorname{dtw}_\gamma \) metric: purple trajectory is the average of red and blue ones.}
    \label{fig:dtw}
\end{figure}

Fig.~\ref{fig:dtw} shows the weighted average $\bar{\mathcal{P}}$ of two primitives $\mathcal{P}_1,\mathcal{P}_2$ with different initial velocities, with $\lambda_1 = \lambda_2 = 0.5$ and $\gamma = 50$.
It can be observed that both the velocity profile and the trajectory agree with the intuition of averaging.

As an addition note, since solving~\eqref{eq:barycenter} is time-consuming and can hardly be completed in real time, it is pre-computed offline and the solutions are stored in a lookup table. This table maps an initial reduced states $(r,\beta,V)$, which are sampled regularly from a fine grid, to corresponding near-feasible trajectories $\bar{\tau}$ generated by solving the optimization problems \eqref{eq:interp_obj}-\eqref{eq:interp_norm} and \eqref{eq:barycenter}.
In online planning and control, the average trajectory \( \bar{\tau} \) in~\eqref{eq:avg_prim} is obtained by looking up for the nearest neighbor of the current reduced state from the table.

\subsubsection{Online tracking controller}
\label{sec:track}

Given a near-feasible drift primitive $\bar{\mathcal{P}}$ for the current initial reduced state, constructed by weighted averaging multiple feasible primitives with various initial reduced states according to Definition~\ref{def:weighted_avg}, a simple feedforward / feedback controller can be designed to track the primitive online.
The controller is based on the intuition that increasing the wheel speed $\omega$ causes both the speed $V$ and the sideslip angle $\beta$ to increase, and that increasing the steering angle $\delta$ causes the yaw rate $r$ to increase in the corresponding direction. Specifically:
\begin{align*}
    & \omega_t = \bar{\omega}_t - k_V(V_t - \bar{V}_t) - k_\beta(| \beta_t | - | \bar{\beta}_t |), \\
    & \delta_t = \bar{\delta}_t - k_r(r_t - \bar{r}_t),
\end{align*}
for any time step $t$, where $(\bar{r}_t, \bar{\beta}_t, \bar{V}_t, \bar{\omega}_t, \bar{\delta}_t)$ are available from the trajectory of $\bar{\mathcal{P}}$, and $k_r, k_\beta, k_V$ are controller parameters.

\section{RESULTS}
\label{sec:result}

\begin{figure}[!htbp]
    \begin{minipage}{0.4\columnwidth}
        \begin{subfigure}{\textwidth}
            \hspace*{-0.3cm}
            \rotatebox{90}{
                \input{figures/2_circle.pgf}}\caption{8-shaped path}
        \end{subfigure}
        \begin{subfigure}{\textwidth}
            \vspace{-0.8cm}
            \rotatebox{30}{
            \hspace{-0.7cm}
            \input{figures/3_circle.pgf}}
            \vspace{-0.7cm}
            \caption{Three-circle path}
        \end{subfigure}
        \begin{subfigure}{\textwidth}
            \input{figures/5_circle.pgf}
            \vspace{-0.65cm}
            \caption{``Olympic Rings'' path}
        \end{subfigure}
    \end{minipage}
    \begin{minipage}{0.6\columnwidth}
        \hspace{-0.2cm}
        \begin{subfigure}{\textwidth}
            \includegraphics{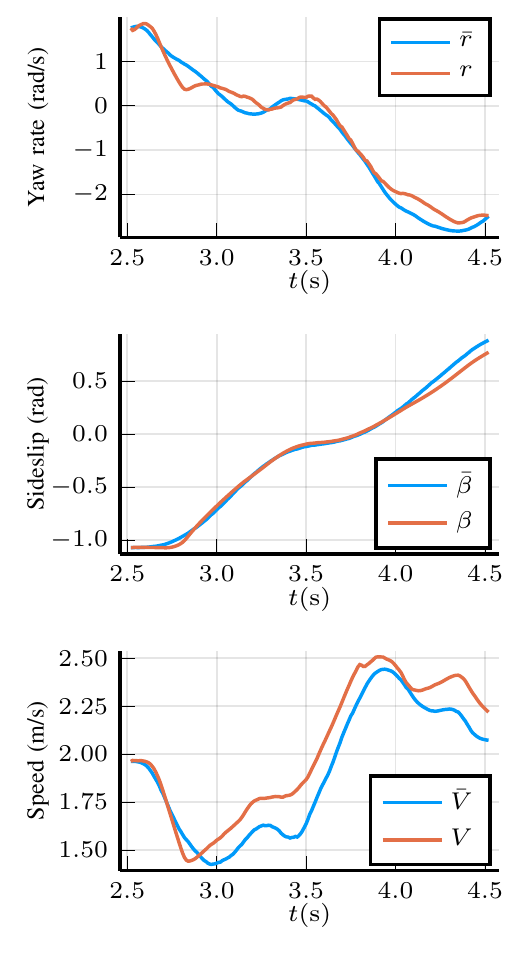}
            \vspace{-0.5cm}
            \caption{Tracking performance}
            \label{fig:tracking}
        \end{subfigure}
    \end{minipage}
    \caption{Simulation result in high-fidelity platform. (a)(b)(c): paths for various tasks (blue for sustained drift, red for inertia drift, gray for target circles); (d): reference and actual reduced state of a single drift primitive ($\bar{r}, \bar{\beta}, \bar{V}$ are available from the reference primitive $\bar{ \mathcal{P}}$).}
    \label{fig:traj}
\end{figure}

In this section, the proposed strategy is verified using the high-fidelity four-wheel simulation and real RC car platforms described in Section~\ref{sec:platform1}. The tire friction characteristics are perturbed in the high-fidelity simulation model to emulate reality gap.

To generate inertia drift primitives from a counterclockwise circular path to a clockwise one, the desired initial reduced state $(r_0, \beta_0, V_0)$ is traversed through the set $\{ 1.5, 1.9, 2.3 \} \times \{ - 0.7, - 1.0, - 1.3 \} \times \{ 1.5, 1.9, 2.3 \}$, where \( \times \) stands for the Cartesian product between sets. This product set covers the range of reduced states that are likely to occur in sustained drifting on the target platform. The desired terminal state $(r_{ \mathrm{des}}, \beta_{ \mathrm{des}}, V_{ \mathrm{des}})$ is fixed at $(-1.9, 1.0, 1.9)$, which corresponds to a clockwise circular path with radius $1 \mathrm{m}$ and a sideslip angle of $1 \mathrm{\ arc}$ on the target platform. For each $(r_0, \beta_0, V_0)$, an inertia drift primitive can be generated in two phases:
\begin{enumerate}
    \item Guide the vehicle starting from rest toward $(r_0, \allowbreak \beta_0, \allowbreak V_0)$ using a linear-quadratic controller based on a locally linearized bicycle model;
    \item Track the ideal inertia drift primitive from $(r_0, \beta_0, V_0)$ to $(r_{ \mathrm{des}}, \beta_{ \mathrm{des}}, V_{ \mathrm{des}})$ obtained by solving~\eqref{eq:opt}, and store the resulting trajectory as a real primitive.
    In particular, we solve~\eqref{eq:opt} using the ALTRO trajectory optimizer~\cite{howell2019altro} and use the tracking controller described in Section~\ref{sec:track}.
\end{enumerate}
Likewise, inertia drift primitives from a clockwise circular path to a counterclockwise one can be generated in a symmetric manner, resulting in a library containing $2 \times 3^3 = 54$ primitives for each target platform.

The performance of the proposed strategy on the high-fidelity simulation platform is shown in Fig.~\ref{fig:traj}. One can observe that the data-driven controller is capable of generating near-feasible reference and delivers a decent tracking performance, and that the primitive-based planner adapts to various tasks through the concatenation of a fixed set of primitives.

\begin{figure}[!htbp]
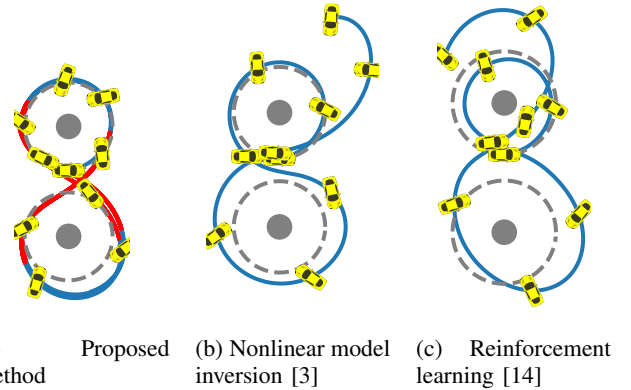

    \vspace{-0.5cm}
    \centering
    \begin{subfigure}{0.3\columnwidth}
        \centering
        \input{figures/2_circle_cmp0.pgf}
        \caption{Proposed method}
        \label{fig:proposed}
    \end{subfigure}
    \;
    \begin{subfigure}{0.3\columnwidth}
        \centering
        \input{figures/2_circle_cmp1.pgf}
        \caption{Nonlinear model inversion~\cite{goh_toward_2020}}
        \label{fig:model_inv}
    \end{subfigure}
    \;
    \begin{subfigure}{0.3\columnwidth}
        \centering
        \input{figures/2_circle_cmp2.pgf}
        \caption{Reinforcement learning~\cite{domberg_deep_2022}}
        \label{fig:rl}
    \end{subfigure}
    \caption{Comparison of trajectories: trajectory is bounded under proposed method, but not under existing methods.}
    \label{fig:compare}
\end{figure}

The proposed method is compared with state-of-the-art methods in autonomous drift~\cite{goh_toward_2020,domberg_deep_2022} using an 8-shaped path as the benchmark problem, which is defined by two circles of \( R = 1 \mathrm{m} \). The results are presented in Fig.~\ref{fig:compare}. As shown in Fig.~\ref{fig:proposed}, the proposed method drives the car through interleaving sustained and inertia drift consecutively, and the trajectories of multiple laps almost overlap. Conversely, as depicted in Fig.~\ref{fig:model_inv} and Fig.~\ref{fig:rl}, the trajectory may diverge under existing methods. The above comparison highlights the advantage of the proposed method over existing ones: the primitive-based planner, a missing component from previous model-based~\cite{goh_toward_2020} and learning-based~\cite{domberg_deep_2022} drift maneuvering methods, suppresses the error accumulated by the low-level controller via the online selection of sustained and inertia drift primitives. Additionally, the data-driven controller's tracking of near-feasible inertia drift trajectories supported by real data demonstrates outstanding performance.

\begin{figure}[!htbp]
    \vspace{0.25cm}
    \includegraphics[width=\columnwidth]{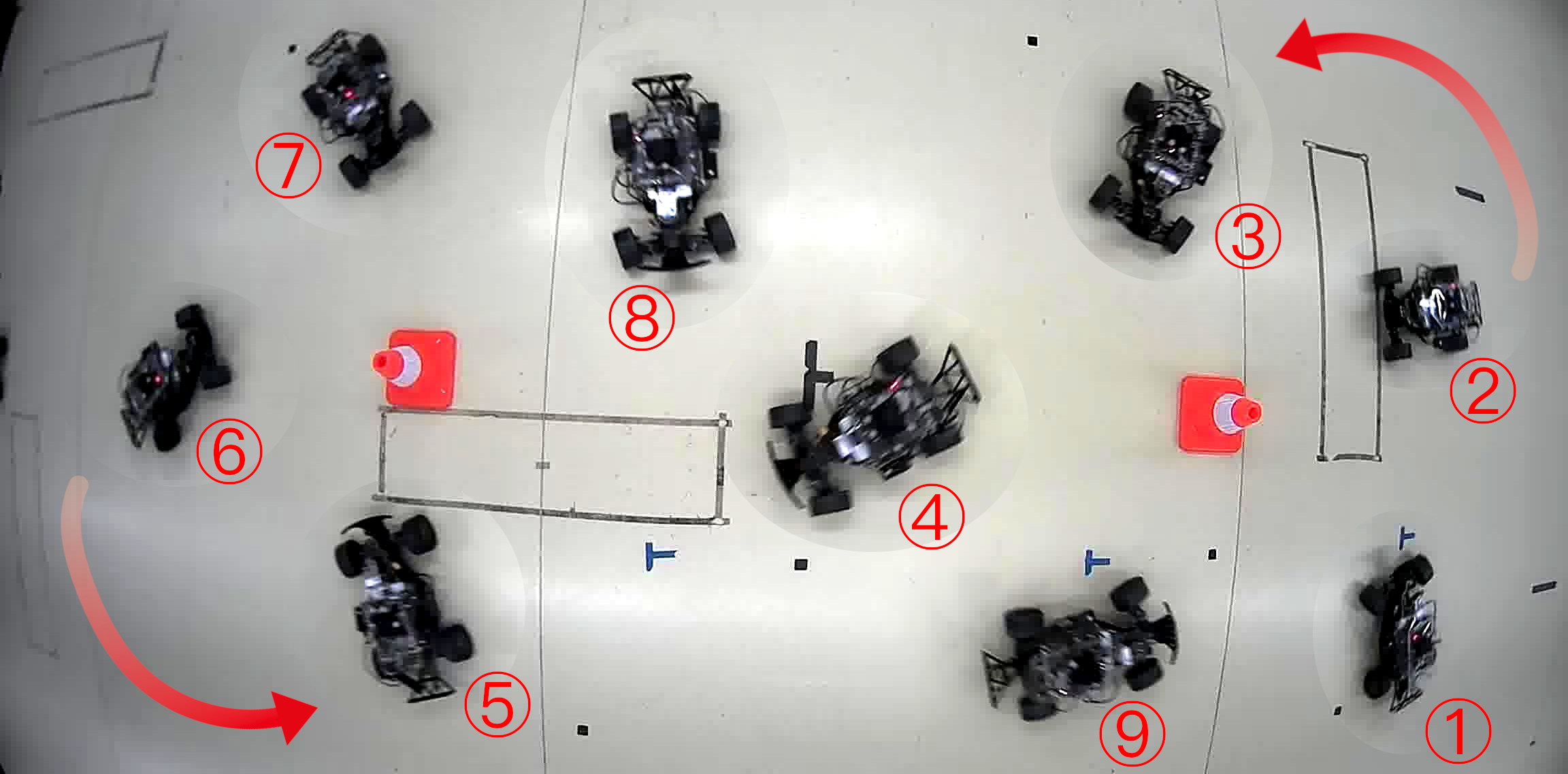}
    \caption{Experiment result for 8-shaped path (9 frames of video stacked together; red digits indicate order of frames)}
    \label{fig:video}
\end{figure}

Snapshots of the hardware experiment are shown in Fig.~\ref{fig:video}, from which one can observe that the proposed strategy maintains consistent performance across simulation and reality. The complete experiment results are included in the accompanying video.

\section{CONCLUSION}
\label{sec:conclusion}

This paper presents a primitive-based planner and data-driven controller for the consecutive execution of inertia drift maneuvers. The proposed strategy is validated in both simulation and reality on a variety of tasks, and shown to outperform existing methods in autonomous drift.
An interesting future direction would be exploring how the inertia drift maneuver can be used in competitive racing settings, and deploying the proposed planning and control strategy to improve racing performance.

\bibliographystyle{IEEEtran}
\bibliography{ref.bib}

\end{document}